\documentclass[letterpaper, 10 pt, conference]{ieeeconf}  

\IEEEoverridecommandlockouts                              

\usepackage{graphics} 
\usepackage{epsfig} 
\usepackage{mathptmx} 
\usepackage{times} 
\usepackage{amsmath} 
\usepackage{amssymb}  
\usepackage{caption}
\usepackage{subcaption}
\usepackage{color}
\usepackage{xcolor}
\usepackage{bookmark}
\usepackage{ifpdf}
\usepackage{cite}
\usepackage{scalerel}
\usepackage{tikz}
\usepackage{makecell}
\usepackage{algorithm}
\usepackage{algorithmic}
\usepackage{stfloats}
\usepackage{float}
\usepackage{multirow}
\usepackage{booktabs}

\title{\LARGE \bf
End-to-End UAV Simulation for Visual SLAM and Navigation
}

\author{Shengyang Chen$^{1}$, 
        Han Chen$^{2}$, 
        Weifeng Zhou$^{1}$, 
        Chih-yung Wen$^{1,2}$, 
        and Boyang Li$^{2}$%
\thanks{$^{1}$Shengyang Chen, Weifeng Zhou, and Chih-yung Wen are with Department of Mechanical Engineering, The Hong Kong Polytechnic University, Hong Kong ({\tt\small shengyang.chen@connect.polyu.edu.hk}).}%
\thanks{$^{2}$Han Chen and Boyang Li are with Interdisciplinary Division of Aeronautical and Aviation Engineering, The Hong Kong Polytechnic University, Hong Kong. }%
}

\usepackage{hyperref}
\hypersetup{
  colorlinks   = true, 
  urlcolor     = blue, 
  linkcolor    = blue, 
  citecolor   = red 
}

\begin{document}

{\textcopyright}IEEE. This paper is a preprint version.

\newpage

\maketitle
\thispagestyle{empty}
\pagestyle{empty}

\begin{abstract}
Visual Simultaneous Localization and Mapping (v-SLAM) and navigation of multirotor Unmanned Aerial Vehicles (UAV) in an unknown environment have grown in popularity for both research and education. However, due to the complex hardware setup, safety precautions, and battery constraints, extensive physical testing can be expensive and time-consuming. As an alternative solution, simulation tools lower the barrier to carry out the algorithm testing and validation before field trials. In this letter, we customize the ROS-Gazebo-PX4 simulator in deep and provide an end-to-end simulation solution for the UAV v-SLAM and navigation study. A set of localization, mapping, and path planning kits were also integrated into the simulation platform. In our simulation, various aspects, including complex environments and onboard sensors, can simultaneously interact with our navigation framework to achieve specific surveillance missions. In this end-to-end simulation, we achieved click and fly level autonomy UAV navigation. The source code is open to the research community.
\end{abstract}

\section*{Supplementary Materials}
Demo video: \url{https://youtu.be/9BQMSVQlo7A} \par

\section{Introduction}
With modern artificial intelligence algorithms, the multi-rotor UAVs are empowered to become smart agents that can navigate in the unknown environment. Given a destination, the UAV can precept the environment, reconstruct the environment map, and dynamically plan a trajectory to the destination. In detail, three sorts of tool kits are applied in such scenarios: localization, mapping, and planning. \par

The localization (or named pose estimation) kit utilizes the onboard sensor information, such as a stereo camera, to estimate the vehicle's 6 degree of freedom (DoF) pose in real-time. The pose feeds into the flight control unit (FCU) to achieve position level control. Given the vehicle pose and sensor input, such as point cloud, the mapping kit reconstructs the environment throughout the mission. Typically, the environment is presented by a 3D occupancy voxel map with the euclidean signed distance information \cite{oleynikova2016signed}. The path planning kit, finds the path to the destination with the minimum cost, avoids the obstacle, and generates a trajectory. The trajectory is then sent to the FCU in the time sequence and navigate the vehicle. \par

Verifying such UAV navigation system under realistic scenarios can be effort-intensive, and a testing failure may lead to damage to the vehicle. The simulator provides simulated hardware components such as perception sensors to researchers. It also helps researchers with the ease of reconfiguration, and the flexibility of environmental setup. \par

\begin{figure}[t!]
     \centering
     \begin{subfigure}[t]{0.45\textwidth}
         \centering
         \includegraphics[width=\textwidth]{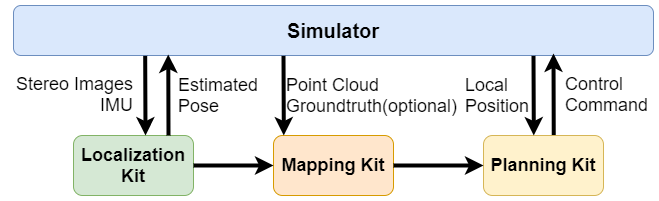}
         \caption{System overview of the proposed simulator.}
         \label{fig:sys_overview}
     \end{subfigure}
     \begin{subfigure}[t]{0.45\textwidth}
         \centering
         \includegraphics[width=\textwidth]{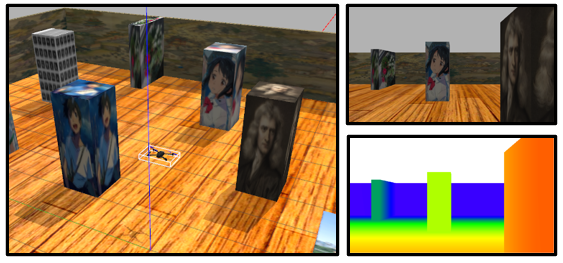}
         \caption{UAV in the simulator, the right hand side images are the real-time color and depth image from the on-board cameras.}
         \label{fig:flyin_gazebo}
     \end{subfigure}
     \begin{subfigure}[t]{0.45\textwidth}
         \centering
         \includegraphics[width=\textwidth]{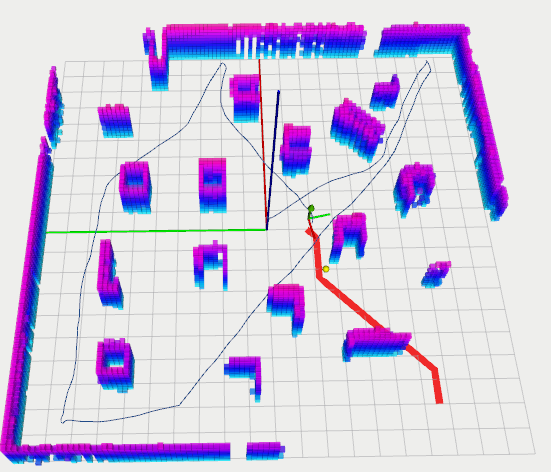}
         \caption{Click and fly navigation in the unknown environment (the blue path is the traveled path and the red path is global planned path from the current position to the destination).}
         \label{fig:clickandfly_rviz}
     \end{subfigure}
        \caption{End-to-end UAV v-SLAM and navigation simulator.}
        \label{fig:overview}
\end{figure}

Various UAV simulation works exist nowadays. However, most of them barely focus on a specific task. The flight dynamic oriented simulation ignore all environment information. The simulation for the v-SLAM study simplified the dynamic model of the vehicle. The navigation-oriented simulation's environment setting contains the 3D information of the environment but neglecting the features of these 3D object. Some simulation works \cite{zhang2015high} \cite{alzugaray2017short} do achieve autonomous navigation in a certain level, but the flexibility is limited and the source codes are not released. The motivation of this work is to construct an end-to-end simulation environment for research and education purposes. Here, the end-to-end refers to the capability of verifying all the perception, reaction, and control algorithm in one simulator (Figure \ref{fig:sys_overview}). \par

Base on the popular ROS-Gazebo-PX4 toolchain, we made several improvements to meet the requirement of UAV v-SLAM and navigation simulation (Figure \ref{fig:flyin_gazebo}). These improvements include: (a) construction of a simulation world, (b) customized UAV models, (c) adding a stereo camera model, and (d) configuration of the vision-based control setup . At the end of this paper, we demonstrate an end-to-end UAV navigation simulation (Figure \ref{fig:clickandfly_rviz}). In summary, the contributions of this work include:
\begin{itemize}
	\item Customization of the ROS-Gazebo-PX4 simulator in terms of the support of stereo inertial vision estimation, vision feedback control, and ground truth level evaluation.
	\item Integration of functions including localization, mapping, and planning into tool kits and achievement of click and fly level autonomy in the simulation environment.
	\item Release of the simulation setup\footnote{https://github.com/HKPolyU-UAV/mav\_sim\_gazebo}, together with the localization kit\footnote{https://github.com/HKPolyU-UAV/FLVIS}, mapping kit\footnote{https://github.com/HKPolyU-UAV/glmapping} and planning kit\footnote{https://bitbucket.org/arclabadmin/fuxi-planner} as open-source tools for the research community.
\end{itemize}

\section{Related Works}

\subsection{UAV Simulator}
We sort the UAV simulator into two categories concerning the scope of simulation: flight dynamics simulator and environment integrated simulator. The first kind of simulator focus on simulating the dynamics for different UAV platforms. All environmental information is neglected except for the gravity force. Based on Matlab Simulink, Quad-Sim \cite{dch33quadsim} is suitable to test the flight control algorithms for different dynamic models. Sun \textit{et al.} \cite{Sun2018} developed another Simulink-based simulator. It includes a comprehensive aerodynamic model of the tail-sitter VTOL UAV. The perception supported simulator, as it named, includes the model of the perception sensors and the environment information. Users can access the simulated sensor output, such as the camera images and the point cloud from the Lidar sensor \cite{Furrer2016}. Schmittle \textit{et al.} \cite{schmittle2018openuav} developed an easy access web-based UAV testbed for education and research. With the Containers as a Service (CaaS) technology, the simulator was deployed on the cloud, which means the user does not need a high-performance computer to execute the simulation.

\subsection{UAV v-SLAM and Navigation System}
Typically, the v-SLAM and navigation system consists localization, mapping, and planning modules, we review the related works in sequence.
\subsubsection{Localization}
Visual localization or visual pose estimation has monocular or stereo camera solutions. The monocular solution has an advantage in terms of simple structure, light-weight, and cost-efficiency. However, recover the scale correctly is the challenge for such a system. Researchers integrate IMU information \cite{qin2018vins} or use predefining the object pattern \cite{frost2018recovering, pfrommer2019tagslam} of the environment to eliminate this problem. Nowadays, stereo camera solutions are off the shelf. As the depth information can be directly extracted from every frame, the accuracy and the robustness of the system are better than the monocular setup. Admittedly, the stereo data stream is more massive than the monocular one. The powerful onboard computers can compensate for this issue. \par

In the UAV application, the visual information is usually fused with the IMU data through either a filter-based framework or an optimization-based framework. In the filter-based framework, the pose and the landmark are in the system states. IMU input propagates the pose states and the relevant con-variance matrix \cite{mourikis2007multi, bloesch2017iterated}. In the optimization-based framework, the IMU engaged through a pre-integration edge \cite{leutenegger2015keyframe}. According to Delmerico \textit{et al.} \cite{delmerico2018benchmark}, the optimization-based approach outperforms the filter-based approach in terms of accuracy but requires more computation resources.

\subsubsection{Mapping}
The mapping system, which provides a foundation for onboard motion planning, is an essential component in the perception-planning-control pipeline. A mapping system needs to balance the measurements' accuracy and the overhead of storage. Three kinds of maps have successful stories in the UAV navigation application: point cloud map \cite{gao2019flying}, occupancy map \cite{hornung2013octomap}, and Euclidean Signed Distance Fields (ESDFs) \cite{oleynikova2017voxblox} map. 

We can easily get the point cloud map by stitching points measurement. However, this kind of map only suitable for high precision sensors in static environments since the sensor noise and dynamic objects cannot be accessed and modified. Occupancy maps, such as Octomap \cite{hornung2013octomap}, store occupancy probabilities in a hierarchical octree structure. These approaches' main restriction is the fixed-size voxel grid, which requires a known map size in advance and cannot be dynamically changed \cite{oleynikova2017voxblox}. Nowadays, Euclidean Signed Distance Fields (ESDFs) map gains popularity \cite{han2019fiesta}. This kind of map is suitable for dynamically growing maps and have the advantage of evaluating the distance and gradient information against obstacles.

\subsubsection{Planning}
For UAV route planning, algorithms can be classified into two main categories, sampling-based \cite{lavalle1998rapidly} and optimization-based \cite{mellinger2011minimum}. Rapidly-exploring random tree (RRT) \cite{lavalle1998rapidly} is the representative of the sampling-based algorithm. In this method, samples are drawn randomly from the configuration space and guide the tree to grow towards the target \cite{gao2017gradient}. Rapidly-exploring random graph (RRG) \cite{karaman2011sampling} is an extension of the RRT algorithm, and it is asymptotic optimal. Even though the sampling-based method is suitable for finding safe paths, it is not smooth for UAV to follow. The minimum snap \cite{mellinger2011minimum} algorithm can be applied to generate a smooth trajectory. It formulates the trajectory generation problem as a quadratic programming (QP) problem. By minimizing the cost function instantly, the trajectory can be represented in piecewise polynomial functions. The cost function includes two terms: the penalty for the trajectory with the potential of collisions and the smoothness of the trajectory itself. In optimization-based methods, another way to add constraints to the optimization problem is first obtaining a series of waypoints by sample search or grid search, then optimize the motion primitives to generate a smooth trajectory through the waypoints under the UAV's dynamic constraints \cite{chen2019dynamic}. It combines the advantages of the two categories, and computation efficiency has a great advantage over those pure optimization-based algorithms. However, the safe radius and other parameters must be tuned carefully. 

\subsection{UAV SLAM and Navigation Simulation} 
Some similar simulations have been conducted prior to this work. Zhang \textit{et al.} \cite{zhang2015high} presented a quadrotor UAV simulator integrated with a hierarchical navigation system. The UAV equips with two laser scanners and one monocular camera. One laser scanners were mounted on the bottom for altitude control and 3D map construction, while another one was attached to the top for navigation purposes. The monocular camera is attached to a tilting mechanism for target detection and vision guidance. The fused data were constructed into Octomap and facilitate a trajectory from an A* global path planner. However, in this work, the environment setup is quite simple and does not include vision features. The navigation is based on the Laser SLAM result. In the simulation from Alzugaray \textit{et al.} \cite{alzugaray2017short}, a point-to-point planner algorithm was designed to work with the SLAM estimation of a monocular-inertial system. The UAV in the simulator fly around the building and reconstruct the environment. However, the UAV is flying outside the building and the obstacle avoidance feature is not considered in this work.

\section{Simulation Platform}
\subsection{Overview}
In the robotic society, the robot operating system (ROS\footnote{https://www.ros.org}) \cite{quigley2009ros} is undoubtedly the most convenient platform, which provides powerful developer tools and software packages from drivers to state-of-the-art algorithms. Besides, many navigation kits have the ROS version package and very convenient to be integrated. Moreover, Gazebo\footnote{http://gazebosim.org} \cite{koenig2004design}, an open-source robotics simulator, is the most widely used simulator in ROS. And we selected the widely used open-source UAV autopilot stack PX4\footnote{https://px4.io} \cite{meier2015px4}. It supports software in the loop (SITL) simulation. Our simulation platform is based on the ROS-Gazebo-PX4 toolchain. \par

\begin{figure}[ht]
\centering
\includegraphics[width=\linewidth]{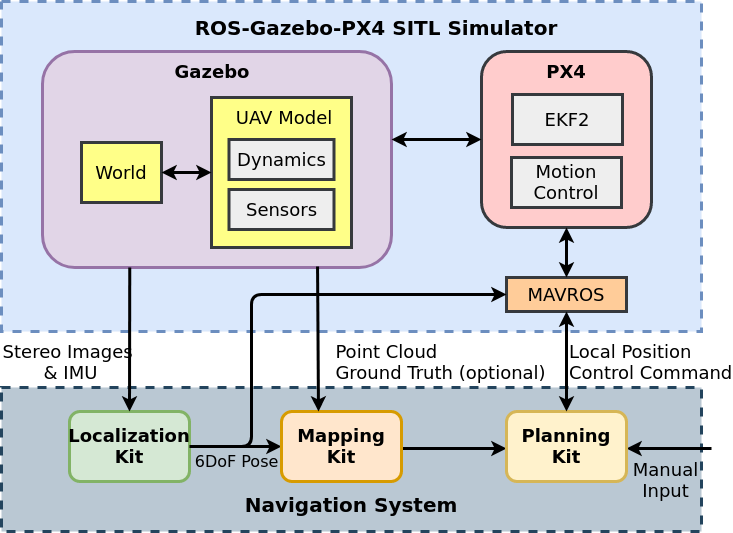}
\caption{Framework of simulation. }
\label{fig:simulation_framework}
\end{figure}

As shown in Figure \ref{fig:simulation_framework}, the upper part is the SITL simulator, and the bottom part is the mapping and navigation system. All components were coordinate through different ROS topics. Especially, the communication between the navigation system and PX4 are through MAVROS\footnote{http://wiki.ros.org/mavros}. We list details of important ROS topics in Appendix. In the simulator, PX4 communicates with the Gazebo to receive sensor data from the simulated world and send the motor and actuator commands back. The Extended Kalman Filter (EKF) based state estimator and motion control module are running on the PX4 stack. In Gazebo, it has an environment map called the world and a simulated UAV model; the UAV model supports the dynamic simulation. Besides, serials of onboard sensors, including GPS, IMU, barometer, and a custom defined depth camera are attached to the UAV model through the Gazebo plugins. \par

\subsection{Dynamics of UAV Model}
The dynamic model in the simulator follows the conventional quadcopter dynamic model which can be found in general papers about dynamics and control such as \cite{lee2010geometric, Mahony2012, Verling_2016}. The UAV is modeled as a 6 DoF rigid body (3 DoF in position,3 DoF in rotation). The position of the center of gravity (C.G.) of the vehicle in the inertial frame is defined by $\pmb{\xi} = [X~Y~Z]^\mathsf T \in \mathbb{R}^3$, the orientation of the vehicle is denoted by the rotation matrix from body frame ($B$) to inertial frame ($I$) $\mathbf{R}_B^I \in SO(3)$, the velocity, as the derivatives of the position, in the inertial frame is described by $\mathbf v = [\dot X~\dot Y~\dot Z]^\mathsf T \in \mathbb{R}^3$, and the angular velocity is denoted by $\pmb \omega$. The kinematics and dynamics of the position and attitude can be denoted by:

\begin{equation}
\label{eq:eq1}
\begin{aligned}
\dot{\pmb{\xi}} &= \mathbf v,\\
m\dot{\mathbf{v}} &= \mathbf{R}_B^I \mathbf{F}_B,\\
\dot{\mathbf{R}}_B^I &=  \mathbf{R}_B^I \pmb \omega_\times,\\
\mathbf{I}\dot{\pmb{\omega}}&=-\pmb{\omega}\times(\mathbf{I}\pmb{\omega})+\mathbf{M}_B,
\end{aligned}  
\end{equation}
where $m$ denotes the mass and $\mathbf I$ denotes the inertia matrix of the vehicle. $\pmb \omega_\times$ denotes the skew-symmetric matrix such that $\pmb \omega_\times \mathbf v = \pmb \omega \times \mathbf v$ for any vector $\mathbf v \in \mathbb{R}^3$. The $\mathbf{F}_B$ and $\mathbf{M}_B$ are total force and moment acting on the body frame, respectively. The dynamic simulation of the UAV is achieved by the model in the Gazebo environment. The PX4 SITL simulator handles all the inner loop attitude, velocity, and position control. The user commands the UAV in the offboard control mode by sending the target position or velocity commands through MAVROS.

\subsection{On-board Sensors}
\begin{figure}[!htb]
    \centering
    \includegraphics[width=0.9\linewidth]{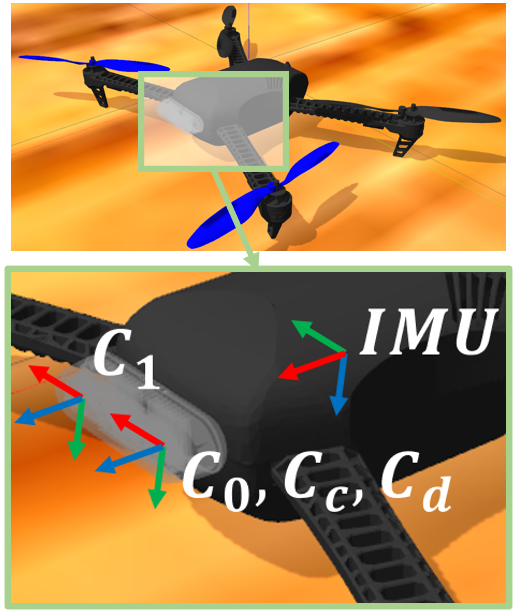}
    \caption{Modified 3DR-IRIS model and the installation geometry of the vision sensor and IMU (X-Y-Z axis in reference frame are colored in red, green, and blue).}
    \label{fig:modified_iris}
\end{figure}

Improved from the 3DR-IRIS model, we added a depth camera and customized the IMU sensor to support the visual-inertial pose estimator. Here we introduce the camera and IMU models. In the Appendix, we introduced how to transplant to other pose estimators. The coordinate definition of the body and IMU is shown in Figure \ref{fig:modified_iris}. \par
\subsubsection{Visual sensor}

The visual sensor is bases on librealsense\_gazebo\_plugin\footnote{https://github.com/pal-robotics/realsense\_gazebo\_plugin}. The visual sensor in the simulator consists of two grey scale cameras ($C_0$ and $C_1$), a color camera ($C_c$), and a depth camera ($C_d$). All these cameras are based on the non-distortion pine hole camera model. The output of the sensor includes two grey scale images, a color image, a depth image, and a point cloud. All these outputs are time synchronized. The horizon field of view ($HFOV$) and resolution ($width \times height$) define the intrinsic camera parameters ($f_x$, $f_y$, $c_x$, $c_y$) by \par

\begin{equation}
f_x=f_y = \frac{width}{2\cdot tan(HFOV/2)}
\label{equ:intrinsic1}
\end{equation}
\begin{equation}
c_x=c_x=\frac{width}{2};c_y=\frac{height}{2}.
\label{equ:intrinsic2}
\end{equation}

Besides, the installation geometry in the SDFormat (Simulation Description Format) file defines the extrinsic parameters. For convenience, the color camera, the depth camera, and the left pine hole camera are installed in the same link, which means the color image, depth image, left grey scale image, and the point cloud are all aligned. \par

\subsubsection{IMU}
In the simulator, the IMU consists of a 3-axis accelerometer and a 3-axis gyroscope. The measured angular velocity and acceleration can be described by the following models:
\begin{align}
&\boldsymbol{\omega}_{m}=\boldsymbol{\omega}_{real} + \boldsymbol{b}_{\omega}+\boldsymbol{n}_{\omega}\\
&\boldsymbol{a}_{m} = \boldsymbol{a}_{real}+\boldsymbol{b}_{a}+\boldsymbol{n}_{a}\\
&\boldsymbol{n}_{\omega} \sim \mathcal{N}(\boldsymbol{0},\boldsymbol{\sigma}_{\omega}^2); \boldsymbol{n}_a \sim \mathcal{N}(\boldsymbol{0},\boldsymbol{\sigma}_a^2)\\
&\boldsymbol{\dot{b}}_\omega \sim \mathcal{N}(\boldsymbol{0},\boldsymbol{\sigma}_{\omega_{b }}^2) ; \boldsymbol{\dot{b}}_a \sim \mathcal{N}(\boldsymbol{0},\boldsymbol{\sigma}_{a_b}^2)
\end{align}
where $\boldsymbol{n}_{\omega}$ and $\boldsymbol{n}_{a}$ refer to the intrinsic noises of the sensor which follow the Gaussian distributions. The biases ($\boldsymbol{\omega}_{b}$ and $\boldsymbol{a}_{b}$) are affected by the temperature and change over time. The slow variations in the sensor biases are modelled with or random walk noise in discrete time. That is, the time derivatives of the biases ($\boldsymbol{\dot{b}}_\omega$ and $\boldsymbol{\dot{b}}_a$) follow the Gaussian distributions. Furthermore, in the IRIS model provided by PX4 Firmware, the update rate of IMU is constrained by the MAVROS. To cross the speed limit, we add another IMU plugin, which publishes the IMU information at the rate of 200 Hz.\par

\subsection{Simulation World Setup}
\begin{figure}[ht]
    \centering
    \includegraphics[width=\linewidth]{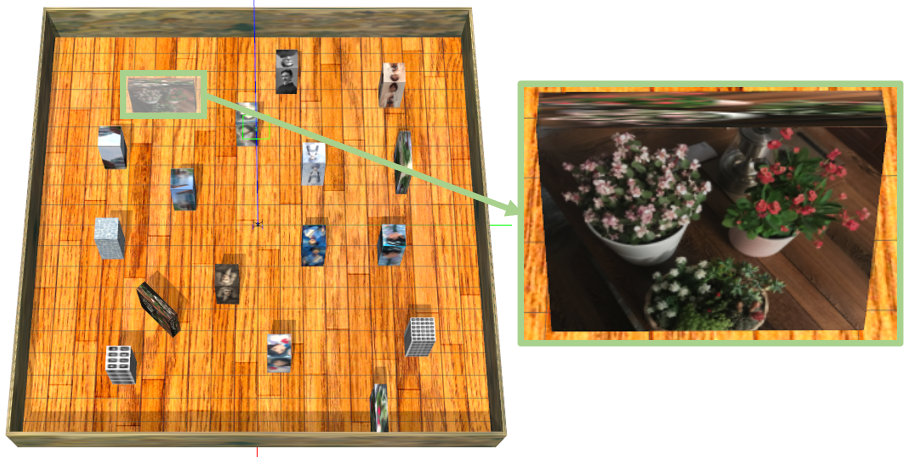}
    \caption{The $20\times20~m$ simulation world with obstacles (wall papers contain rich visual features to support the visual tracking).}
    \label{fig:fig4}
\end{figure}

First, we added obstacles such as walls and boxes into a $20\times20~m$ empty world. Then, to meet the requirement of v-SLAM simulation, we furnish all these items and the ground plane with the wallpapers, which contains rich visual features, shown in Figure \ref{fig:fig4}. 

\section{MAV Navigation Framework}

\subsection{Localization}

\begin{figure}[!htb]
    \centering
    \includegraphics[width=\linewidth]{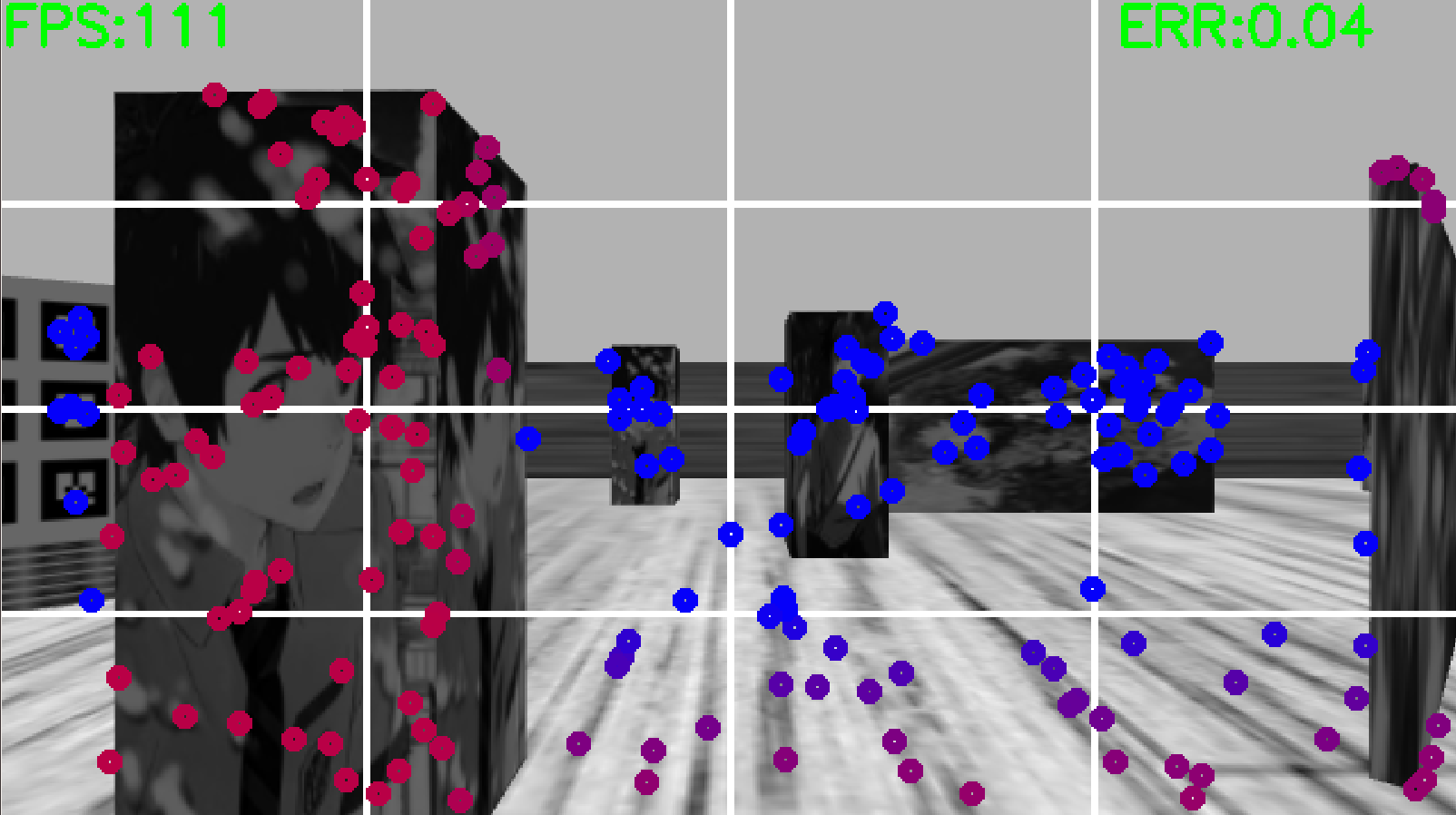}
    \caption{FLVIS in the simulation, the markers in the image refer to the landmarks, different color refer to the distance between the landmark and the camera.}
    \label{fig:flvis}
\end{figure}

We integrate FLVIS\cite{chen2020stereo}, a stereo visual-inertial pose estimator developed by our group, as the localization kits (Figure \ref{fig:flvis}). Compared to other monocular vSLAM solutions, stereo visual-inertial pose estimator has the advantages of robustness, accuracy, and scale-consistency. They could be further explained as Robustness: the pose between consecutive visual frames can be estimated by IMU when visual tracking is lost. Accuracy:  more measurements are fused in the pose estimation process to get better accuracy. Scale-consistency: the depth information can be extracted directly from stereo images without any motion. FLVIS use the feedback/feedforward loops to fuse the data from IMU and stereo/RGB-D camera and achieve high accuracy on resource-limited computation platform.

\subsection{Mapping}

\begin{figure}[!htb]
    \centering
    \includegraphics[width=\linewidth]{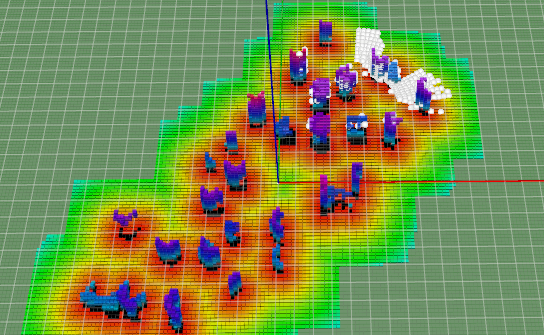}
    \caption{Visualization of Global/Local/ESFDs map in Rviz. In the global map, the color (blue-purple) refers to the height of the obstacle; the ESFDs map's color (red-yellow-green) refers to the signed distance value; the highlight white spheres refer to the local map.}
    \label{fig:glmapping}
\end{figure}

A global-local mapping kit `glmapping' was integrated into the simulator as shown in Figure \ref{fig:glmapping}. This mapping kit is a 3D occupancy voxel map designed for the MAV or mobile robot navigation applications. Currently, most of the navigation strategies are the combination of global planning and local planning algorithms. Global planning focuses on finding the least cost path from the current position to the destination. Furthermore, local planning is used to replan the trajectory to avoid obstacles. This mapping kit processes the perception information separately. The global map, on the cartesian coordinate system, is a probability occupancy map. The local map, on cylindrical coordinates system, has an excellent dynamic performance. The mapping kit also supports the projected 2D occupancy grid map and ESDFs map output.

\subsection{Path Planning}

\begin{algorithm}[b]
\caption{fuxi-planner} 
\label{alg1}
\begin{algorithmic}[1]
\WHILE{goal not reached:} 
\STATE Receive a global 3D voxel map ${Pcl_{m}}$, and its projecttion on the ground (${Map1}$) as the 2D pixel map for path finding
\STATE Cut off blank edge of ${Map1}$ and apply obstacle inflation on ${Map1}$, output ${Map2}$
\STATE Find the shortest ${Path1}$ to goal
\STATE Calculate the optimal local goal by the Bezier curve
\ENDWHILE
\WHILE{goal not reached:}
\STATE Receive the local goal
\STATE Find the next waypoint $N_k$ by heuristic angular search

\IF{${found a feasible waypoint:}$} 
\STATE Run the minimum acceleration motion planner to get motion primitives
\ELSE 
\STATE Run the backup plan for safety, then go to 5
\ENDIF
\STATE Send the motion primitives to the UAV flight controller
\ENDWHILE
\end{algorithmic}
\end{algorithm}

We integrated fuxi-Planner \cite{chen2019dynamic, chen2020computationally} as our path planning kit. Algorithm \ref{alg1} illustrates the planning process. The planner is composed of a global path planner and a local planner. The global planner works on a 2D global grid map to find the shortest 2D path and output the local goal for the local planner. And the local planner works directly on point cloud to avoid the potential collision with obstacles and plan a kinematically feasible trajectory to the local goal. The local planner's core component is a sample-based waypoint search method called the heuristic angular search (HAS) method. The engagement of the global planner prevents the local planner from failing into the kidnap situation. The global planner utilizes Jump Point Search (JPS) algorithm \cite{harabor2014improving} to output a serial of waypoints, which present the shortest path. The first three waypoints are used as the control points to draw a Bezier curve. The local path planner's goal is to locate at the tangent at the Bezier curve's first waypoint. The two planners work in parallel. Although the global planner's outer loop frequency is relatively low, the local planner inner loop still maintains a high update rate and can continuously command the UAV.

\section{Results and Discussion}
In this section, we conducted two experiments to show to performance of the proposed simulator. In the first experiment, we flew the vehicle manually in the simulation world with the keyboard to verify the localization and mapping kits. In the second experiment, we demonstrated the click and fly level autonomous navigation. \par

\subsection{Manual Exploration}
In this case, we integrated the localization kit and mapping kit. With the first person view (FPV) from the colored camera and the real-time reconstructed map view, we controlled the UAV to explore the $20 m \times20 m$ unknown environment. This exploration mission cost 7 minutes and 24 seconds, and the UAV traveled 82 meters in the simulation world. \par

\begin{figure}[!htb]
    \centering
    \includegraphics[width=\linewidth]{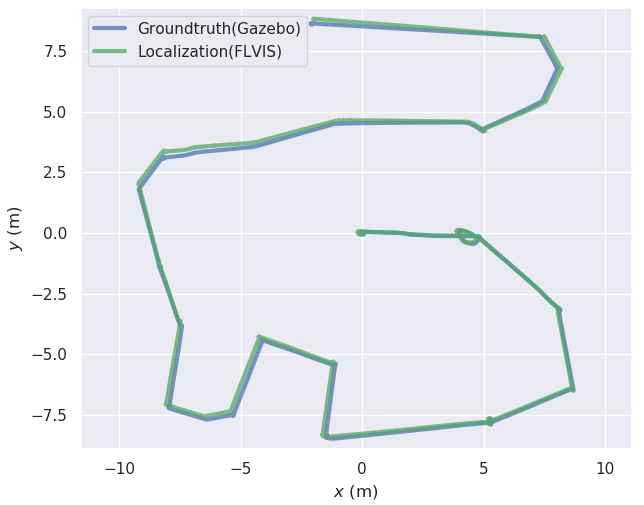}
    \caption{Comparison of the ground truth and the estimated pose from the localization kit.}
    \label{fig:manual_explore_path}
\end{figure}

The performance of the localization and mapping kit were then evaluated. The excellent agreement between the ground truth path and the estimated path form the localization kit can be observed in Figure \ref{fig:manual_explore_path}. We used a tool from Michael Grupp \cite{grupp2017evo} to evaluate the accuracy of the localization kit. The absolute trajectory error (ATE) of the translation drift in the form of root mean square error (RMSE) is 0.3 m. 

\begin{figure}[!htb]
    \centering
    \includegraphics[width=\linewidth]{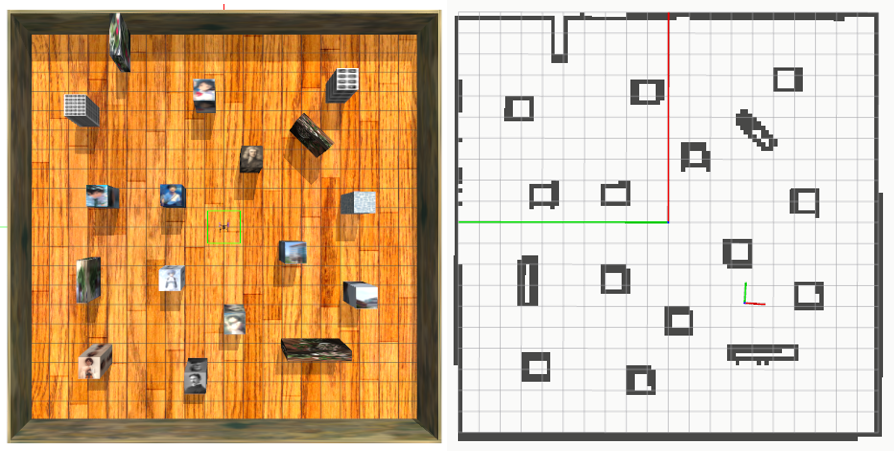}
    \caption{Top-down view of the simulation world (left) and the projected 2D occupancy grid map generated by the mapping kit (right).}
    \label{fig:img_topdown}
\end{figure}

\begin{figure}[!htb]
    \centering
    \includegraphics[width=\linewidth]{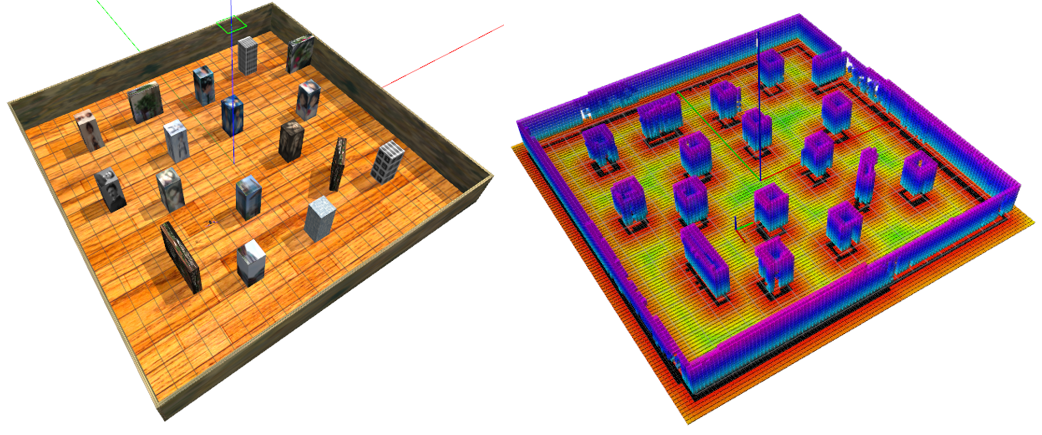}
    \caption{Oblique view of the simulation world (left) and reconstructed map by the mapping kit (right).}
    \label{fig:img_oblique_view}
\end{figure}

We capture the image of the simulation world and reconstructed map from different views. Good agreement between them and the detail of the map can be observed in Figure \ref{fig:img_topdown} and \ref{fig:img_oblique_view}. The voxel size of the map is $0.2 \times 0.2 \times0.2$ meter.


\subsection{Click and Fly Level Autonomy}

\begin{figure}[!htb]
    \centering
    \includegraphics[width=\linewidth]{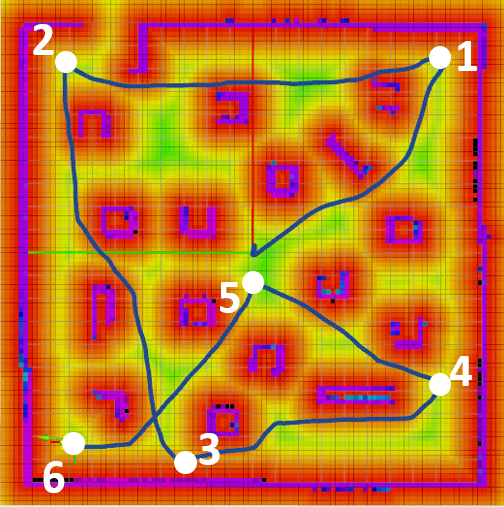}
    \caption{Click and fly navigation (the blue path is the traveled path and the white mark in the image is the given way point).}
    \label{fig:caf_map}
\end{figure}

In this case, we further integrated the path planning kit into the simulation. We only give a desired destination on the map to the UAV. Then the UAV will plan a path, avoid the obstacle, and fly to the destination automatically. As shown in Figure \ref{fig:caf_map}, six waypoints were set during the mission. The drone precept the environment, plan a path to visit these waypoints in sequence fully automatically. The fly path kept a safe distance to the nearest obstacle to avoid the collision.
\par

\begin{figure}[!htb]
    \centering
    \includegraphics[width=\linewidth]{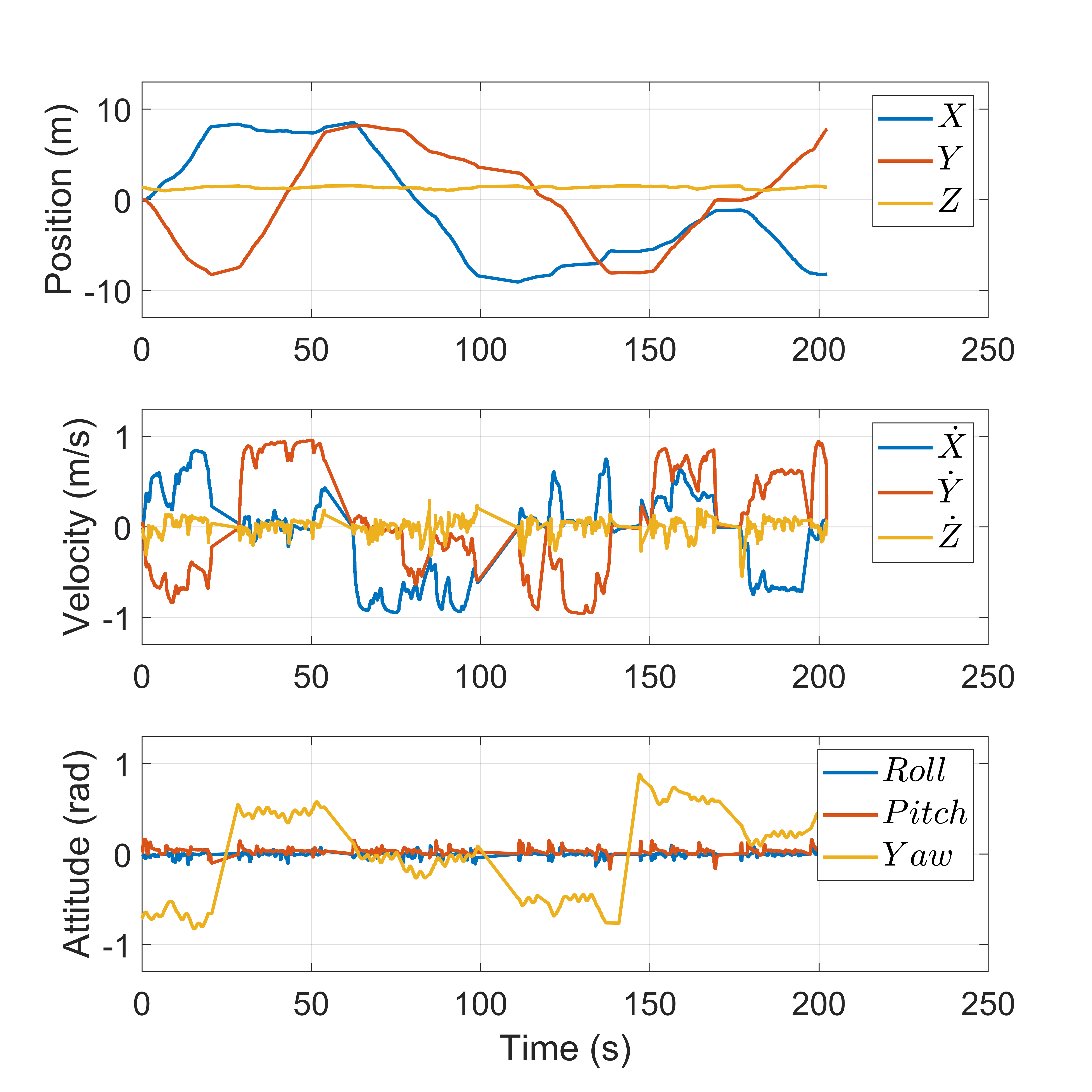}
    \caption{The record of position, velocity, and attitude in click and fly navigation.}
    \label{fig:caf_avp}
\end{figure}

Figure \ref{fig:caf_avp} displays the curve of position, velocity, and attitude of the simulation. We can see the velocity is well controlled under the max speed limit (1 m/s) and the position curves are fluent and pass all the six goal points stably. The pitch and roll angles are small so that the vision localization kit work at high performance.\par

\setcounter{table}{0}
\renewcommand{\thetable}{A\arabic{table}}
\begin{table*}[bp] 
\centering
\caption{IMPORTANT ROS TOPICS}
\begin{tabular}{llll}
\toprule
\textbf{Topic name}                                                & \textbf{Topic type}                 & \textbf{Update rate(Hz)} & \textbf{Comment}                              \\ \hline
\multicolumn{4}{l}{\textbf{Simulator}}                                                                                                             \\ \hline
/camera/infra1/image\_raw                                 & sensor\_msgs/Image         & 30          & Image of the left grey scale camera  \\ \hline
/camera/infra2/image\_raw                                 & sensor\_msgs/Image         & 30          & Image of the right grey scale camera \\ \hline
/camera/color/image\_raw                                  & sensor\_msgs/Image         & 30          & Image of the color camera            \\ \hline
/camera/depth\_aligned\_to\_color\_and\_infra1/image\_raw & sensor\_msgs/Image         & 30          & Depth Image                          \\ \hline
/camera/depth/color/points                                & sensor\_msgs/PointCloud2   & 30          & Point Cloud (aligned to color)        \\ \hline
/iris/imu                                                 & sensor\_msgs/Imu           & 200         & IMU data                             \\ \hline
/gt\_iris\_base\_link\_imu                                & nav\_msgs/Odometry         & 50          & Ground truth of the IMU link         \\ \hline
\multicolumn{4}{l}{\textbf{with Localization Kit}}                                                                                                 \\ \hline
/imu\_path                                                & nav\_msgs/Path             & 200         & Path at IMU rate                     \\ \hline
/vision\_path                                             & nav\_msgs/Path             & 30          & Path at vision rate                  \\ \hline
/vo\_img0                                                 & sensor\_msgs/Image         & 30          & Output image with colored landmanks  \\ \hline
/vo\_img1                                                 & sensor\_msgs/Image         & 30          & Colored depth Image                  \\ \hline
\multicolumn{4}{l}{\textbf{with Mapping kit}}                                                                                                      \\ \hline
/globalmap                                                & visualization\_msgs/Marker & 10          & Visualize global map                 \\ \hline
/localmap                                                 & visualization\_msgs/Marker & 10          & Visualize local map                  \\ \hline
/occupancygrid                                            & nav\_msgs/OccupancyGrid    & 30          & Projected occupancy grid map         \\ \hline
/esfd\_map                                                & visualization\_msgs/Marker & 10          & Visualized ESFD map                  \\ \hline
\multicolumn{4}{l}{\textbf{with Planning kit}}                                                                                                     \\ \hline
/jps\_path                                                & nav\_msgs/Path             & 10-20           & Global path on grid map                         \\ \hline
/local\_wp                                                & visualization\_msgs/Marker             & 50-70           & waypoint of the local planner                         \\ \hline
/global \_goal                       & geometry\_msgs/Point             & 10-60           & Temporary goal for local planner                                      \\ \hline
/mavros/setpoint\_velocity/cmd\_vel                       & geometry\_msgs/PoseStamped             & 50-70           & Velocity command for PX4 controller                      \\ \bottomrule
\end{tabular}
\label{tab:rostopics}
\end{table*}

\subsection{Performance Analysis}
The configurations for the navigation system in the simulation are as follows: in localization kit, the input image resolution is $640 \times 360$; in mapping kit, the voxel size is $0.2m \times 0.2m \times 0.2m$ and the map contains 181500 ($110 \times 110 \times 15$) voxels. The simulation has been verified on two different computers. The processing times are listed in the Table \ref{tab:processing_time}. The average time factor refer to the ratio of the real time over the simulation time. The value of time factor is 1 means the simulation runs in real-time.

\setcounter{table}{0}
\renewcommand{\thetable}{\arabic{table}}    
\begin{table}[H] 
\centering
\caption{PROCESSING TIME OF THE NAVIGATION SYSTEM AND THE SIMULATION TIME FACTOR}
\begin{tabular}{llcc}
\toprule
\multicolumn{2}{c}{\textbf{Kits}}                                  & \textbf{Computer 1} & \textbf{Computer 2} \\ \hline
\multicolumn{2}{l}{Localization (with out loop clousure)} & 28 ms      & 22 ms      \\ \hline
\multirow{3}{*}{Mapping}        & Global map                & 26 ms      & 18 ms      \\ \cline{2-4} 
                                & Local map                 & 4 ms       & 4 ms       \\ \cline{2-4} 
                                & Projected ESDFs map       & 70 ms      & 64ms        \\ \hline
\multirow{2}{*}{Planning}       & Global planning           & 90 ms      & 65ms        \\ \cline{2-4} 
                                & Local planning            & 20 ms      & 16ms        \\ \hline
Simulator                       & Average time factor       & 0.6        & 0.92       \\ \bottomrule
\end{tabular}
\vspace{1ex}
	
{\raggedright Note: Computer 1: i5-8250u CPU, 8GB RAM, GeForce MX150 graphic card; Computer 2: i7-8550u CPU, 16GB RAM, GeForce MX150 graphic card.\par}
\label{tab:processing_time}
\end{table}

\section{CONCLUSIONS AND FUTURE WORK}
In this letter, we introduced an end-to-end UAV simulation platform for SLAM and navigation research and applications, with the detailed simulator setup and an the out of box localization, mapping, and navigation system. We also demonstrated the click and fly level autonomy navigation by the simulator. The flight results show that the simulator could provide trustworthy data stream and also the versatile interfaces for autonomous functions development. We offered all the kits for public access to promote further research and development of the autonomous UAV system based on this framework.Future work will focus on two aspects. One is to support more open source notable navigation-related kits. Moreover, to design the benchmark scenario in the simulator to evaluate the performance of these kits. Another aspect is to expand the current simulator to encompass more perception sensors, more UAV platforms, and more challenging environments for a variety of potential tasks. \par




\section*{APPENDIX}
\label{appendix:importantrostopics}
\subsection{Important ROS Topics}
Several important ROS topics are list in Table \ref{tab:rostopics}.

\subsection{Support for Stereo Visual Inertial Pose Estimator}
In default simulator setup. The resolution of the camera is $640\times360$ with the $HFOV=1.5~rad$. According to Equation \ref{equ:intrinsic1} and \ref{equ:intrinsic2}, the camera intrinsic is
$$[f_x,f_y,c_x,c_y]=[343.4963, 343.4963, 320, 180].$$
The extrinsic parameter including the transformation from right camera to left camera $\mathbf{T}_{C_1}^{C_0}$ and the transformation from left camera to IMU $\mathbf{T}_{C_0}^{I}$. By the default installation geometry setup, they are:
$$
\mathbf{T}_{C_1}^{C_0}=\begin{bmatrix}
1 & 0 & 0 & 0.05\\ 
0 & 1 & 0 & 0\\ 
0 & 0 & 1 & 0\\ 
0 & 0 & 0 & 1
\end{bmatrix},
\mathbf{T}_{C_0}^{I}=\begin{bmatrix}
0  & 0  & 1 & 0.12\\ 
-1 & 0  & 0 & 0\\ 
0  & -1 & 0 & 0\\ 
0  & 0  & 0 & 1
\end{bmatrix}.$$

\bibliographystyle{ieeetr}
\bibliography{ref}

\end{document}